\pdfoutput=1

\documentclass[11pt]{article}

\usepackage{authblk}

\usepackage[]{ACL2023}

\usepackage{times}
\usepackage{latexsym}

\usepackage[T1]{fontenc}

\usepackage[utf8]{inputenc}

\usepackage{microtype}

\usepackage{inconsolata}
\usepackage{paralist}
\usepackage{graphicx}
\usepackage{mdframed}

\usepackage{url}
\usepackage{xspace}
\usepackage{mathtools}
\usepackage{multirow}
\usepackage{microtype}
\usepackage{pifont}
\usepackage{pgfplots}
\pgfplotsset{compat=1.8}
\usepackage{relsize}
\usepackage{subcaption}
\usepackage{tikz}
\usepackage{tabularx}

\def\impressions{\textsc{impressions}\xspace}
\def\findings{\textsc{findings}\xspace}
\def\mimicthree{\textsc{MIMIC-III}\xspace}
\def\mimicfour{\textsc{MIMIC-IV}\xspace}
\def\mimiccxr{\textsc{MIMIC-CXR}\xspace}
\def\chexpert{\textsc{CheXpert}\xspace}

\def\eqref#1{Eq.~\ref{eqn:#1}}

\def\tabref#1{Table~\ref{tab:#1}}
\def\tablabel#1{\label{tab:#1}\label{p:#1}}

\newcommand\blfootnote[1]{%
  \begingroup
  \renewcommand\thefootnote{}\footnote{#1}%
  \addtocounter{footnote}{-1}%
  \endgroup
}
\title{shs-nlp at RadSum23: Domain-Adaptive Pre-training of Instruction-tuned LLMs for Radiology Report Impression Generation}

\author[1]{\bf Sanjeev Kumar Karn$^\ast$}
\author[1]{\bf Rikhiya Ghosh$^\ast$}
\author[1]{\bf Kusuma P$^\ast$}
\author[1]{\bf Oladimeji Farri}
\affil[1]{Digital Technology and Innovation, Siemens Healthineers}
\affil[ ]{\tt \{sanjeev.kumar\_karn,rikhiya.ghosh,kusuma.p,oladimeji.farri\}@siemens-healthineers.com}

\begin{document}
\maketitle
\blfootnote{$^\ast$ Equal Contribution.}
\begin{abstract}
Instruction-tuned generative Large language models (LLMs) like ChatGPT and Bloomz possess excellent generalization abilities, but they face limitations in understanding radiology reports, particularly in the task of generating the \impressions section from the \findings section. They tend to generate either verbose or incomplete \impressions, mainly due to insufficient exposure to medical text data during training. We present a system which leverages large-scale medical text data for domain-adaptive pre-training of instruction-tuned LLMs to enhance its medical knowledge and performance on specific medical tasks. We show that this system 
performs better in a zero-shot setting than a number of \textit{pretrain-and-finetune} adaptation methods on the \impressions generation task, and ranks 1st among participating systems in Task 1B: Radiology Report Summarization at the BioNLP 2023 workshop.   

\end{abstract}

\section{Introduction}
A radiology report is the primary means by which the radiologist communicates his/her interpretations of medical images (e.g., X-Rays) and resulting conclusions to the ordering physician. A radiology report typically includes several sections \cite{kahn2009toward}, among which the most important ones are the \findings and \impressions sections. \findings includes qualitative and quantitative descriptions of abnormalities if present, along with the radiologist's diagnosis or differential diagnosis regarding the observations. \impressions summarises the \findings section, and the radiologist notes major abnormalities and their recommendations in this subsection; a sample report with \findings and \impressions is shown in \tabref{report_sample}. 
There have been various efforts to automatically generate \impressions from \findings, such as, \newcite{karn-etal-2022-differentiable} reinforcement learning- and \newcite{DelbrouckRadSum23} BERT-based systems.  

Pretrained language models (PLMs) are trained on enormous, heterogeneous corpora that include everything from news articles and literary works to web content and encyclopedia articles, which allows them to capture a broad range of linguistic patterns and features \cite{gururangan2020don}. However, PLMs also have limitations and biases, particularly in their training data, which can affect their performance in certain out-of-domain tasks \cite{bommasani2021opportunities}. To advance the state of the art, several domain-specific PLMs are available, specifically in the field of biomedical and clinical NLP, such as BioBERT \cite{lee2020biobert} and RadBERT \cite{yan2022radbert}. These PLMs do, however, have additional constraints. For instance, the datasets used to train RadBERT were rather small and only covered a few number of anatomical specialities.

\begin{table}[!t]
\begin{center}
\begin{small}
\begin{tabular}{m{0.98\linewidth}}
\hline
\multicolumn{1}{c}{\findings}\\
\hline
bifrontal hemorrhagic contusions are once again noted, stable compared to most recent prior, \ldots. subarachnoid hemorrhage is once again noted within the left \ldots. subdural hematoma is noted overlying the left temporal lobe and to the left \ldots. there is no shift of normally midline structures. the ventricles appear unremarkable. a left temporal lobe hemorrhagic contusion remains stable in size \ldots. the visualized paranasal sinuses are clear. there is no evidence of acute fracture.
\\
\hline
\multicolumn{1}{c}{\impressions}\\
\hline
 1. bifrontal hemorrhagic contusion appears stable compared to most recent prior with slightly increased vasogenic \ldots\\
 2. subdural hematoma is noted layering over the left temporal lobe and within the left falx. \\
 3. subarachnoid hemorrhage is noted within the left frontal region. \\
 4. no shift of normally midline structures.\\
\hline
\end{tabular}
\end{small}
\end{center}
\caption{\tablabel{report_sample}
\findings (top) and \impressions (bottom) sections of a radiologist's report from \mimicthree.} 
\end{table}

The \textit{pretrain-and-finetune} paradigm for PLMs has become the dominant approach for addressing downstream tasks with significant training data scarcity \cite{karn-etal-2021-shot}. 
Recent studies (e.g., \newcite{gururangan2020don}) suggest that additional pretraining on in-domain text, referred to as specialist pretraining, prove more successful for improving downstream performance. 
Currently another approach has emerged, where instead of finetuning PLMs to perform downstream tasks, the objectives of downstream tasks are reconstructed using textual prompts similar to the original pre-training objectives \cite{liu2023pre}. This \textit{pretrain-and-prompt-tune} paradigm is commonly referred to as prompt-tuning.

Multitask prompted finetuning (also known as instruction tuning) is a type of large-scale \textit{pretrain-and-prompt-tune} paradigm where finetuning of large PLMs (also referred as LLMs) is performed with datasets representing various NLP tasks defined by instructions as natural language prompts \cite{scao2022bloom}. 
Using this approach, \newcite{scao2022bloom} equipped their LLM Bloom with the skill to perform multilingual zero-shot instruction-based tasks and tagged it Bloomz.

We propose an extension of the domain adaptation paradigm beyond the typical method of \textit{pretrain-and-finetune} or instruction-tuned LLM for domain-specific tasks.
We posit that additional pretraining of LLMs that already went through the vast \textit{pretrain-and-prompt-tune} with in-domain text adapts better and more easily. We refer to our approach as \textit{general-pretrain-prompt-tune-and-special-pretrain}. With this approach, the model is trained using the same initial LM objective in each of the three training stages (i.e., general pretraining, prompt-tuning and domain specialized pretraining), which is a significant advantage. Furthermore, since the instruction-tuned LLM is familiar with a variety of prompts and has tackled numerous tasks, the domain pretraining saturates rapidly, resulting in lower training costs. 

We continued the pretraining of the instruction-tuned Bloomz on \mimicfour to form RadBloomz, and evaluated this adaptation paradigm on the radiology report summarization task. 
The proposed system in a zero-shot setting exhibits better performance than \textit{pretrain-and-finetune} methods, and ranks 1st place among participating systems in Task 1B: Radiology Report Summarization at the BioNLP 2023 workshop. 

Overall, our contributions are as follows:
\begin{itemize}
    \item We extend the domain adaptation paradigm by introducing \textit{general-pretrain-prompt-tune-and-special-pretrain} by further pretraining instruction-tuned LLMs like Bloomz on the domain-specific text.
    \item We show that the new adaptation paradigm of an instruction-tuned LLM for radiology yields better performance in a zero-shot setting than \textit{pretrain-and-finetune} methods.
\end{itemize}


\section{Datasets}
\subsection{Pretraining Datasets for Radiology Domain Adaptation.}
We have performed domain-adaptive pretraining using the recently published \mimicfour radiology reports dataset \citep{johnsonmimic, physiobank2000physionet}. It contains over 2.3 million radiology reports from 237k patients, and approximates to over 616 million tokens with Bloomz \citep{muennighoff2022crosslingual} tokenizer. After preprocessing, we only used 1.4 million reports with 190 million tokens. The \tabref{lang-diff} provides further details on the statistics of the datasets.

\subsection{Finetuning Datasets for Impression Generation}
We utilize the datasets that were shared for Task 1B: Radiology Report Summarization at the BioNLP 2023 workshop for our finetuning task. The task consists of three datasets: \mimicthree \citep{johnson2016mimic}, \mimiccxr \citep{johnson2019mimic} and \chexpert \citep{irvin2019chexpert}, pre-split into \findings and \impressions sections. For \mimicthree, there are 59320 reports in training dataset, 7413 in validation and 6526 in test set and 6531 in hidden test set. Most reports (91.4\%) pertain to CT imaging, and the most represented anatomy pertains to the head (52.8\%). Although the task related to \mimiccxr/\chexpert datasets is multimodal, we only used radiology reports for finetuning and inference. \mimiccxr training dataset consists of 125,417 radiology reports in training dataset, 991 in validation and 1624 in test dataset. The hidden test dataset is a \chexpert dataset and consists of 1k reports. 

\begin{table}[t]
\footnotesize
\centering
\begin{tabular}{|l@{\hspace{0.08cm}}|r|r|}
\hline
 Dataset &\findings & \impressions \\
 \hline
\multirow{1}*{\mimicfour} &  113.46(139.06) & 33.04 (36.52) \\
\hline
\multirow{1}*{\mimicthree} & 118.19(59.7) & 49.48 (35.12) \\
\hline
\multirow{1}*{\mimiccxr} & 54.52 (24.67) & 16.37 (15.79) \\
\hline
\end{tabular}
\caption{\tablabel{lang-diff}
Number of words per report with standard deviation in parentheses for various dataset. 
}
\end{table}

\section{Methods}
Our methods consists of preprocessing, domain-adaptive pretraining, finetuning and inference. 

\subsection{Preprocessing}
The preprocessing step uses Regex-based cleaning and normalization to remove spurious characters and administrative or repetitive texts unrelated to the report. We have added special tokens for de-identified text in the reports. In addition, we also identified different sections in the report, namely \findings and \impressions. We have selected the reports with less than 512 tokens that have both these sections.

\subsection{Domain adaptive pretraining (DAPT)}
We select GPT-powered Bloom \cite{scao2022bloom} as the base LLM for our study. Bloom is a multilingual LLM modified from Megatron-LM GPT2 \citep{shoeybi2019megatron} trained auto-regressively with an objective function of cross entropy with mean reduction. There are multiple versions of Bloom based on the number of parameters. The largest Bloom model consists of 176 billion parameters, with 70 layers, 112 attention heads and 14336 dimensional hidden layers. Bloomz \citep{muennighoff2022crosslingual} is a massive multitask instruction-tuned version of Bloom. For our domain adaptation study, we use a variant of Bloomz (Bloomz-7b1) with 7 billion parameters, 30 layers and 4096-dimensional hidden layers.

Following our proposed \textit{pretrain-fine-tune-and-pretrain} paradigm, we continuously pretrain Bloomz-7b1 using cross-entropy loss on auto-regressively generated tokens from the \findings and \impressions sections of \mimicfour reports. 

\subsection{Finetuning}
In this study, the domain-specific task for finetuning an LLM is Radiology Report Summarization. Following the standard prompt-based finetuning, we use \findings as the prompt and finetune Bloomz-7b1 using cross-entropy loss of the auto-regressively generated summary tokens by comparing with the ground-truth \impressions. We also appended \textsc{TL;DR} to the prompt. This technique keeps the final fine-tuning objective consistent with the pretraining and instruction-tuning objectives of the base Bloom and intermediate Bloomz. In order to avoid catastrophic forgetting, we have reduced trainable parameters of Bloomz-7b1 by freezing all but the last layer of the model.

\subsection{Inference}
The inference pipeline leverages the trained model to generate \impressions, given the \findings. The metrics used to evaluate the generated results are Rouge scores \cite{lin2004rouge}, F1RadGraph \citep{delbrouck2022improving}, Bertscore \citep{zhang2019bertscore} and F1CheXbert \cite{xie2023factreranker} for \mimiccxr/\chexpert datasets.

\begin{table*}
\footnotesize
\centering

\resizebox{0.90\textwidth}{!}{
\begin{tabular}{p{0.15\textwidth}|c|ccccc}
\textbf{Team}&\textbf{hidden test\-set}&\textbf{BLEU4} & \textbf{ROUGE-L} & \textbf{BertScore}& \textbf{F1-cheXbert}&\textbf{F1-RadGraph}\\
\hline
\bf shs-nlp&\multirow{5}*{\mimicthree} &\bf 18.36 &\bf35.32& \bf57.26	& N/A &	\bf36.94\\
utsa-nlp& &16.05 &34.41& 57.08	& N/A &	36.31\\
aimi& &16.61 &33.43& 55.54	& N/A &	35.12\\
sinai& &17.38	&32.32	&55.04	&N/A	&33.96\\
knowlab& &13.23	&32.02	&55.64	&N/A	&33.39\\
\hline
dmis-msra&\multirow{5}*{\mimiccxr} &\bf18.62&	\bf34.57&	\bf55.90&	\bf72.36&	\bf43.20\\
utsa-nlp& &16.33	&34.97	&55.54	&69.41	&42.86\\
knowlab& &14.41	&33.63	&54.72	&67.20	&39.98\\
\bf shs-nlp& &14.59	&32.43	&53.99	&68.99	&38.40\\
aimi& &5.15	&31.84	&47.83	&64.18	&32.05\\
\end{tabular}
}
\caption{\tablabel{mimic-hidden-result}
The table shows performance of top-5 submitted systems on the two categories of hidden test data of the shared task 1B at BioNLP 2023. shs-nlp is our RadBloomz system. Hidden test-set \mimicthree includes only reports while \mimiccxr includes reports and images. Our system is text-based and thus \mimicthree is more appropriate evaluation and ranks 1st among participating systems.  
}
\end{table*}

\begin{table*}
\footnotesize
\centering

\resizebox{0.90\textwidth}{!}{
\begin{tabular}{p{0.2\textwidth}|c|ccccc}
\textbf{Models}&\textbf{open test-set}& \textbf{BLEU4} & \textbf{ROUGE-L} & \textbf{BertScore}& \textbf{F1-cheXbert}&\textbf{F1-RadGraph}\\

\hline
RBz-0shot&\multirow{2}*{\bf MIMIC-III} 
&\bf17.33& 33.93&	55.49&	N/A&	\bf34.93\\
RBz-ft && 16.49 & \bf35.25 & \bf57.29 & N/A& 31.12\\
\hline
RBz-0shot&\multirow{2}*{\bf MIMIC-CXR} &\bf25.32&	\bf47.48&	\bf63.61&	\bf74.34&	\bf49.00\\
RBz-ft &&  16.16 & 26.16 & 52.22 & 53.1 & 31.07\\
\end{tabular}
}
\caption{\tablabel{mimic-result}
Results for different domain adaptation on the different test split of the shared task 1B. The experimental setup is the same for all methods, i.e., the same train/validation/test split of the medical reports was used. RBz-0shot: RadBloomz-zero shot, RBz-ft: RadBloomz-finetuned. 
}
\end{table*}

\section{Experiments}
We propose two experimental runs for the summarization task. 
 \begin{enumerate}
     \item \textbf{Radiology Domain Adaptive Pretraining (RadBloomz) with \mimicfour and zero-shot inference.} GPT-powered Bloomz-7b1 model is further trained using causal language objective on \mimicfour radiology report to form RadBloomz. We set the sequence length to 512, training batch size to 64, validation batch size to 32, learning rate to 3e-5 and AdamW as the optimizer \citep{loshchilov2017decoupled}. The best zero-shot inference results are for 24k steps.
     \item \textbf{RadBloomz finetuned with \mimicthree.} We follow \textit{pretrain-and-finetune} paradigm and finetuned the RadBloomz further with \mimicthree dataset on radiology report summarisation task. We use the same hyperparameters and training configuration as the above. The best finetuning results are for 2697 steps.
 \end{enumerate}
All the experiments were run on the same infrastructure. \footnote{Eight Tesla A100 SXM4 GPUs (with 80 GB memory per GPU) using Deepspeed zero-3 configuration \citep{rasley2020deepspeed} with BF16 enabled} We use sampling based technique to generate the summary from the model output distribution given \findings. We set sampling hyper-parameter such as maximum tokens to 128, $top\_k$ to 50 and $top\_p$ to 0.7. 

\section{Results and Discussion}
We compared results from RadBloomz to various systems using n-gram overlap ROUGE and facts overlap F1RadGraph evaluations.

\tabref{mimic-hidden-result} highlights the performance of RadBloomz (with team name shs-nlp) on \mimiccxr and \mimicthree hidden test datasets. Hidden test-set \mimicthree includes only reports while \mimiccxr includes reports and images. Our system is text-based only and thus \mimicthree is a more appropriate evaluation. Our system is the top performer for \mimicthree hidden test set among all the other submitted systems. Additionally, for \mimiccxr hidden test set our text-based system ranks fourth among all the other submitted systems, further showing the strength of the proposed domain adaptation technique on a multi-modal task.   

In \tabref{mimic-result}, we compare the performance of the standard \textit{pretrain-and-finetune} and our proposed paradigm \textit{pretrain-prompt-tune-pretrain-and-zero-shot} on the radiology report summarization test data for the Task 1B challenge.\footnote{Ground-truths are only available for the open test data for any additional evaluation.} We note that although finetuning with \mimicthree improves the Rouge-L and Bertscore metrics for \mimicthree test dataset, the Bertscore, F1-RadGraph and F1-cheXbert scores are lower for the finetuned model. This shows that the domain adaptation paradigm is sufficient for achieving higher performance and doesn't require task-specific finetuning.  

\textbf{Error Analysis.}
A detailed error analysis on the open test datasets reveals that many of the generated impressions get a low score for both Rouge and F1-RadGraph in cases where the radiology report has no abnormalities mentioned. For example, generated impression ``normal mri of the cervical spine.'' and ground truth impression ``negative study'' are semantically similar, but these n-gram overlap-based scores fail to recognize the semantic relatedness. Similarly, we noticed that similar \findings sometimes generate different \impressions. For example, impressions can be as detailed as : ``near complete opacification of the ethmoid air cells and sphenoid sinuses. moderate air-fluid level with mucosal thickening of the right maxillary sinus and moderate mucosal thickening of the left maxillary sinus.'', while similar \findings in another report would be summarised to ``pansinusitis, as described above." In addition, we have noticed problems with missed facts and hallucinations, as seen in \ref{sec:appendix}.

\section{Conclusion}
In this work, we introduce a new domain adaptation paradigm of \textit{general-pretrain-prompt-tune-and-special-pretrain} where we further pretrain an instruction-tuned LLM (Bloomz) on the radiology domain text. We use radiology report summarization as the domain-specific task and demonstrate that the new paradigm-based LLM model performs better than the standard \textit{pretrain-and-finetune} based method even in a zero-shot setting. The system ranks 1st among participating systems in the hidden-test category in Task 1B: Radiology Report Summarization at the BioNLP 2023 workshop.

\section*{Limitations}
There are a few limitations pertaining to the training data we used. Some of them are listed below.
\begin{enumerate}
    \item Our domain Adaptation of LLMs was performed on English reports only, and therefore may not work out of the box in a multilingual setting.
    \item There is data imbalance with respect to imaging modalities and anatomies covered by our training data. For example, regions like  extremities, neck, spine and shoulder are underrepresented in the dataset, and report summarization related to those regions needs to be thoroughly evaluated.
    \item There needs to be a study on the diversity of the patients represented in the data, and how it impacts the performance of the model for underrepresented communities. 
    \item Different radiologists (and radiology departments) have different preferences and styles of writing reports. In addition, clinical referrals sometimes dictate to what extent some details are documented in the report. There was no study on the consistency, uncertainty or information richness of the report.
\end{enumerate}
Asides from the training data, there may be space and time throughputs of the model which could make them unsuitable for on-premise and/or at-the-edge applications. This aspect offers an opportunity for further work on how best to quantize and deploy RadBloomz (and similar LLMs) within the clinical workflow towards improved efficiency for radiologists.

\section*{Ethics Statement}
The research performed in this paper adheres to the Association for Computing Machinery (ACM) Code of Ethical and Professional Conduct \footnote{https://www.acm.org/code-of-ethics} adopted by the Association for Computational Linguistics (ACL). To prevent any harm caused due to errors in our model-generated outputs, our models are meant to be deployed in a human-in-the-loop setting where the key information extracted by our models are reviewed by radiologists and physicians.

\section*{Disclaimer}
The concepts and information presented in this paper/presentation are based on research results that are not commercially available. Future commercial availability cannot be guaranteed.

\bibliography{anthology,impressionLLM}
\bibliographystyle{acl_natbib}
\section{Appendix}
\label{sec:appendix}
\subsection{Hallucinations and missed facts by RadBloomz}

A detailed error analysis has revealed several interesting types of hallucinations by RadBloomz on the MIMIC-III test dataset.

\textbf{Numerical hallucination.}
RadBloomz model has been seen to hallucinate, especially if numbers are involved. There is a problem comparing numbers or relating mentioned numbers to their correct concepts. One example is seen in table \tabref{numerical_hallucination_example} where the model has problem understanding the largest size of the lymph node conglomerate mentioned in the text, though it is explicitly mentioned which one is the largest conglomerate.
\begin{table}[ht!]
\begin{center}
\begin{small}
\begin{tabular}{m{0.96\linewidth}}
\hline
\multicolumn{1}{c}{\findings}\\
\hline
\ldots there is massive lymphadenopathy in the mesentery and retroperitoneum with significant interval worsening compared to prior scan. there are conglomerates of lymph nodes, the largest at the paraaortic region measures 6.7 x 4.4 cm, (2:73). there is a large conglomerate to the right common iliac artery, measuring 4.7 x 4.6 cm, (2:91). \ldots
\\
\hline
\multicolumn{1}{c}{Generated \impressions}\\
\hline
  interval worsening of the massive lymphadenopathy in the mesentery and retroperitoneum, with conglomerates of lymph nodes, largest at the right common iliac artery, measuring 4.7 x 4.6 cm.\\
\hline
\end{tabular}
\end{small}
\end{center}
\caption{\tablabel{numerical_hallucination_example}
Numerical hallucination example: \findings (top) and \impressions (bottom) sections of a radiologist's report from \mimicthree test dataset.} 
\end{table}

\textbf{Missing context and facts.}
There are findings for which the RadBloomz has missed facts or did not understand the context of a sentence well. One such example is shown in \tabref{missing}. 

\textbf{Hallucination of new facts}
New concepts, unrelated to what is mentioned in the findings is a part of the summary. These are rare occurrences. We found only 6 cases out of the 6.5k where these occured. One such example is shown in \tabref{wrong_concept}

\begin{table}[ht!]
\begin{center}
\begin{small}
\begin{tabular}{m{0.9\linewidth}}
\hline
\multicolumn{1}{c}{\findings}\\
\hline
\ldots there is massive inflammatory stranding surrounding the pancreas. no definite fluid collections are seen. \ldots  there is a moderate amount of free fluid seen in the abdomen in the left paracolic gutter. this demonstrates attenuation values of simple fluid. there is also some fluid around the spleen. \ldots
\\
\hline
\multicolumn{1}{c}{Generated \impressions}\\
\hline
 \ldots 2. extensive inflammatory stranding surrounding the pancreas. 3. no definite fluid collections are seen. \ldots\\
\hline
\end{tabular}

\caption{\tablabel{missing}
Missing context and facts : \findings (top) and \impressions (bottom) sections of a radiologist's report from \mimicthree test dataset.} 
\bigskip
\begin{tabular}{m{0.9\linewidth}}
\hline
\multicolumn{1}{c}{\findings}\\
\hline
moderate right and small left pleural effusions have increased from prior examination. there is associated compressive atelectasis. the heart is normal in size, without pericardial effusion. note is made of a small sliding hiatal hernia.
\\
\hline
\multicolumn{1}{c}{Generated \impressions}\\
\hline
 \ldots no evidence of recurrent or metastatic disease in the abdomen or pelvis\ldots\\
\hline
\end{tabular}
\end{small}
\end{center}
\caption{\tablabel{wrong_concept}
Wrong concept hallucination: \findings (top) and \impressions (bottom) sections of a radiologist's report from \mimicthree test dataset.} 
\end{table}

\end{document}